\documentclass[10pt,twocolumn,letterpaper]{article}
\usepackage{authblk} 

\usepackage{cvpr}              
\usepackage{placeins}

\usepackage[dvipsnames]{xcolor}

\usepackage{bm}

\def\vv{{\bm{v}}}

\def\vx{{\bm{x}}}
\def\vy{{\bm{y}}}
\def\vz{{\bm{z}}}

\DeclareMathAlphabet{\mathsfit}{\encodingdefault}{\sfdefault}{m}{sl}
\SetMathAlphabet{\mathsfit}{bold}{\encodingdefault}{\sfdefault}{bx}{n}

\usepackage{cancel}

\definecolor{cvprblue}{rgb}{0.21,0.49,0.74}
\usepackage[pagebackref,breaklinks,colorlinks,citecolor=cvprblue]{hyperref}

\def\confName{CVPR}
\def\confYear{2024}

\title{Adapting Self-Supervised Learning for Computational Pathology}

\author[1]{Eric Zimmermann}
\author[1]{Neil Tenenholtz}
\author[1]{James Hall}
\author[2]{George Shaikovski}
\author[2]{Michal Zelechowski}
\author[2]{Adam Casson}
\author[2]{Fausto Milletari}
\author[2]{Julian Viret}
\author[2]{Eugene Vorontsov}
\author[2]{Siqi Liu}
\author[1,\textdagger]{Kristen Severson}
\affil[1]{Microsoft Research, Cambridge, MA United States}
\affil[2]{Paige, NYC, NY United States}

\begin{document}
\maketitle

\def\thefootnote{\textdaggerdbl}\footnotetext{Presented at DCA in MI Workshop, CVPR 2024}\def\thefootnote{\arabic{footnote}}
\def\thefootnote{\textdagger}\footnotetext{ Corresponding author: kseverson@microsoft.com}\def\thefootnote{\arabic{footnote}}

\begin{abstract}
Self-supervised learning (SSL) has emerged as a key technique for training networks that can generalize well to diverse tasks without task-specific supervision. This property makes SSL desirable for computational pathology, the study of digitized images of tissues, as there are many target applications and often limited labeled training samples. However, SSL algorithms and models have been primarily developed in the field of natural images and whether their performance can be improved by adaptation to particular domains remains an open question. In this work, we present an investigation of modifications to SSL for pathology data, specifically focusing on the DINOv2 algorithm. We propose alternative augmentations, regularization functions, and position encodings motivated by the characteristics of pathology images. We evaluate the impact of these changes on several benchmarks to demonstrate the value of tailored approaches.
\end{abstract}    
\section{Introduction}\label{sec:intro}
Pathology is the study of the causes and effects of disease and is used as a foundation for diagnosis and treatment. Pathology relies on the study of stained tissue microscopy slides, also known as whole slide images (WSIs). Computational pathology (CPath) refers to the application of artificial intelligence to WSIs which have been digitized enabling automated study and characterization~\cite{deng2020deep,cooper2023machine,song2023artificial,srinidhi2021deep}. Several factors have recently converged to drive increased interest in the use of foundation models, large-scale deep neural networks trained on expansive datasets without access to task-specific labels, in CPath. Foundation models have demonstrated incredible success in the natural image domain, even in cases with limited labeled data for training, by learning image representations, called embeddings, which can be used as inputs to a wide-variety of downstream tasks. These properties are desirable for the pathology domain as there are many tasks, such as diagnosis, disease subtyping, biomarker quantification, estimation of treatment response, and survival prediction, and curated labeled datasets are expensive to gather as they require expert review. Motivated by these factors, there have been recent efforts to collect large pathology image datasets and subsequently several foundation models have been proposed \cite{wang2022transformer,ciga2022self,azizi2023robust,kang2023benchmarking,chen2024towards,vorontsov2024virchow,dippel2024rudolfv,filiot2023scaling}. 

The rise of foundation models has been enabled in part by the use of self-supervised learning approaches. Self-supervision is a learning paradigm where a pre-text task is constructed without access to target labels. The quality of features learned and their ability to transfer to a target task is highly dependent on the definition of the pre-text task and its correlation to downstream objectives. Joint-embedding self-supervised learning (JE-SSL) methods pose the learning objective in terms of alignment and diversity~\cite{bordes2023democratizing, wang2022understanding}. Alignment is accomplished by encouraging the embeddings of pairs or sets of samples generated from the same source image via the application of an augmentation policy to be close to one another~\cite{chen2020simple, zbontar2021barlow, bardes2022vicreg}. Diversity provides the necessary support to learn representation that avoid collapse or trivial solutions across the entire set of observations. Examples of diversity objectives include using explicit constraints such as decorrelation, maximum entropy, and minimum energy as well as implicit constraints such as model asymmetry or centering operations \cite{wang2022understanding, chen2020exploring, caron2021emerging, grill2020bootstrap}.
\begin{figure*}[ht]
    \centering
    \includegraphics[width=0.9\textwidth]{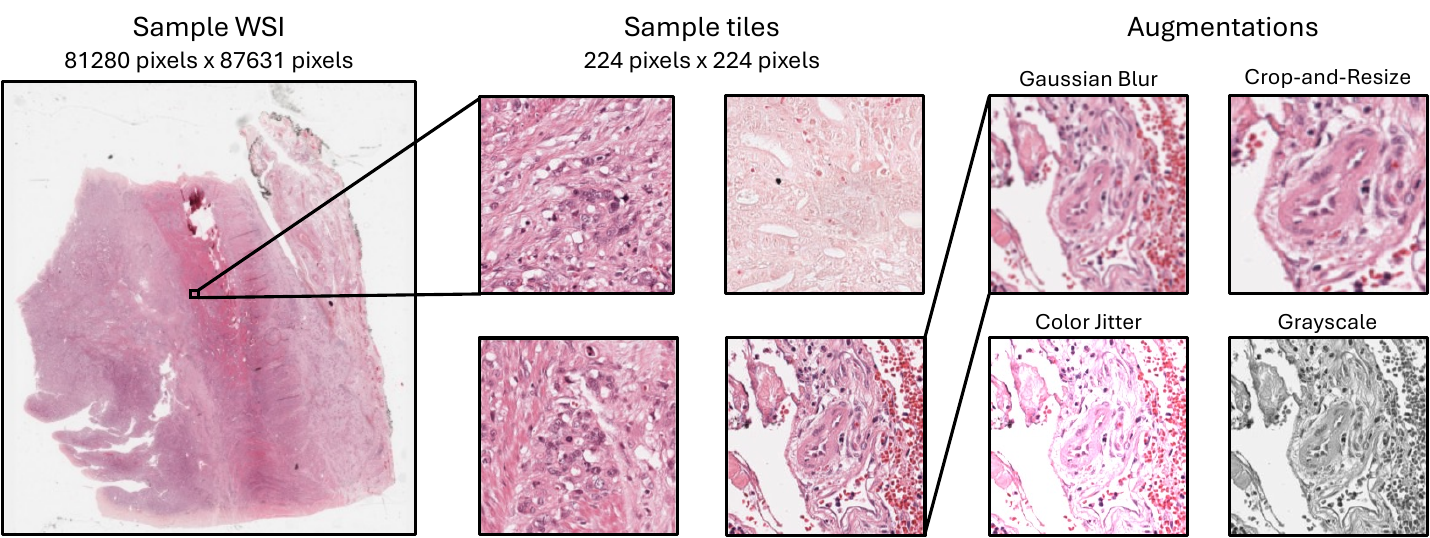}
    \caption{Example of a computation pathology data pipeline for self-supervised learning. Tiles (center) that satisfy a tissue inclusion criteria are randomly sampled from a WSI (left) and perturbed using a set of augmentations sampled from a policy to build the self-supervised pre-text task (right). Augmentations include but are not limited to Gaussian blur, color jitter, grayscale, and crop-and-resize.}
    \label{fig:augmentations}
\end{figure*}

The choice of augmentation policy is a key factor in determining what kinds of features may be learned during training. Often, augmentation strategies designed for natural images are directly applied to other domains, as has largely been the case in computational pathology. However it is unclear that the same augmentations will produce good features across domains. Indeed it has been shown that the effects of augmentation are often data dependent and vary across classes \cite{balestriero2022effects}. Unlike natural images, which are a 2D projection of a 3D scene, CPath images are acquired from essentially 2D slices of tissue that are stained and scanned at various resolutions or magnifications through a digitization process. One implication of this is that pixels have a known corresponding physical distance. Understanding the relative distances and shapes of morphological features may be instructive when interpreting the image. A further distinction of pathology data as compared to natural images is that tissue patterns are non-unique and less diverse across a population. These facts motivate the investigation of changes to standard self-supervised learning strategies.
\FloatBarrier
In this work, we investigate the use of domain-inspired modifications to self-supervised learning algorithms and the ViT architecture focusing on preserving and accounting for morphological patterns. We choose DINOv2~\cite{oquab2024dinov2}, a self-supervised distillation method, as the framework to test the impact of different design choices and evaluate performance on various benchmark tasks. Our results indicate that preserving local morphology and modifying regularization schemes lead to better performance.

\section{Related Work}
\label{sec:related_work}

Possibly in part due to the computational cost, few works have studied the impact of augmentation policies in CPath foundation models on downstream task performance, however there are a few notable exceptions. Tellez et al.~\cite{tellez2019quantifying}, Ciga et al.~\cite{ciga2022self}, and Gullapally et al.~\cite{gullapally2023synthetic} all investigated the impact of color augmentation motivated by stain variation, the effect of differences in staining protocol and scanner type which do not reflect underlying differences in pathology. All of the aforementioned studies have demonstrated improvements in performance using color augmentation and several models have employed domain-orientated approaches~\cite{wang2022transformer,dippel2024rudolfv}. Digitized WSIs are typically stored at fixed magnifications, e.g. 5$\times$, 10$\times$, 20$\times$, and have significantly less variability in object scale than natural images. As noted in the introduction, this feature also draws into question augmentation protocols that typically crop and resize images. In addition to color variation, Ciga et al.~\cite{ciga2022self} investigated the impact of random cropping on model performance and found less random cropping generally improved performance although the largest observed improvement in any setting was 5\%. 

\section{Learning on gigapixel pathology images}\label{sec:mixmag}
WSIs are gigapixel images, typically tens of thousands of pixels in each spatial dimension. In order to avoid the computational bottlenecks associated with large images, foundation models have been trained using local regions of the image called tiles (see Fig.\ref{fig:augmentations}). Tiles are extracted by sub-diving an image into non-overlapping regions that have sufficient amounts of relevant tissue as determined by a separate inclusion criterion such as a segmentation network. To-date, most works have generated tiles at a fixed magnification, typically 20$\times$. As pathology features manifest at various resolutions, we instead consider a set-up where tiles are drawn from the most commonly available magnifications, $5\times$, $10\times$, $20\times$, and 40$\times$, or 2.0, 1.0, 0.5 and 0.25 microns-per-pixel, respectively.

Given this setup, in the next two sections, we discuss how pairwise alignment tasks and diversity objectives can be adapted to be more suited to pathology image features. We focus our description to the DINOv2 algorithm~\cite{oquab2024dinov2} and vision transformer architectures~\cite{dosovitskiy2020image}. The DINOv2 approach uses self-distillation cross-entropy, masked image modeling, and differential entropy to learn image features. It has recently been used to train several CPath foundation models~\cite{dippel2024rudolfv, chen2024towards, vorontsov2024virchow,kang2023benchmarking} and we therefore use it to frame our analysis, however we expect that many of the insights can generalize to other self-supervised learning settings.

\subsection{Morphology-preserving alignment} \label{sec:morphology}
The alignment component of JE-SSL objectives is achieved via an invariance objective that is controlled through a careful selection of an augmentation policy.
The augmentation policy is applied as follows: for a sample $\vx$ drawn from a dataset $\mathcal{D}$, $T$ random perturbations are sampled from the augmentation policy $\mathcal{T}$ and applied to create views $\{\mathcal{T}_t(\vx)\}_{t=1}^T$. These views then serve as input to the alignment objective. Examples of common augmentations are shown in Fig.~\ref{fig:augmentations}.

The design of the augmentation policy determines the properties of the network and must be difficult enough to learn meaningful information without shortcuts. Traditional augmentation policies were designed for object-centric natural images which have common reoccurring structure and texture. Color perturbation and random crop-and-resize are the fundamental building blocks of these policies and aim to decouple global-local features. The motivation for crop-and-resize is the creation of feature co-occurances resulting from overlaps between views which can then be used for alignment~\cite{han2022augmentation,huang2021towards}.

\begin{figure}[h]
    \centering
    \includegraphics[width=0.95\columnwidth]{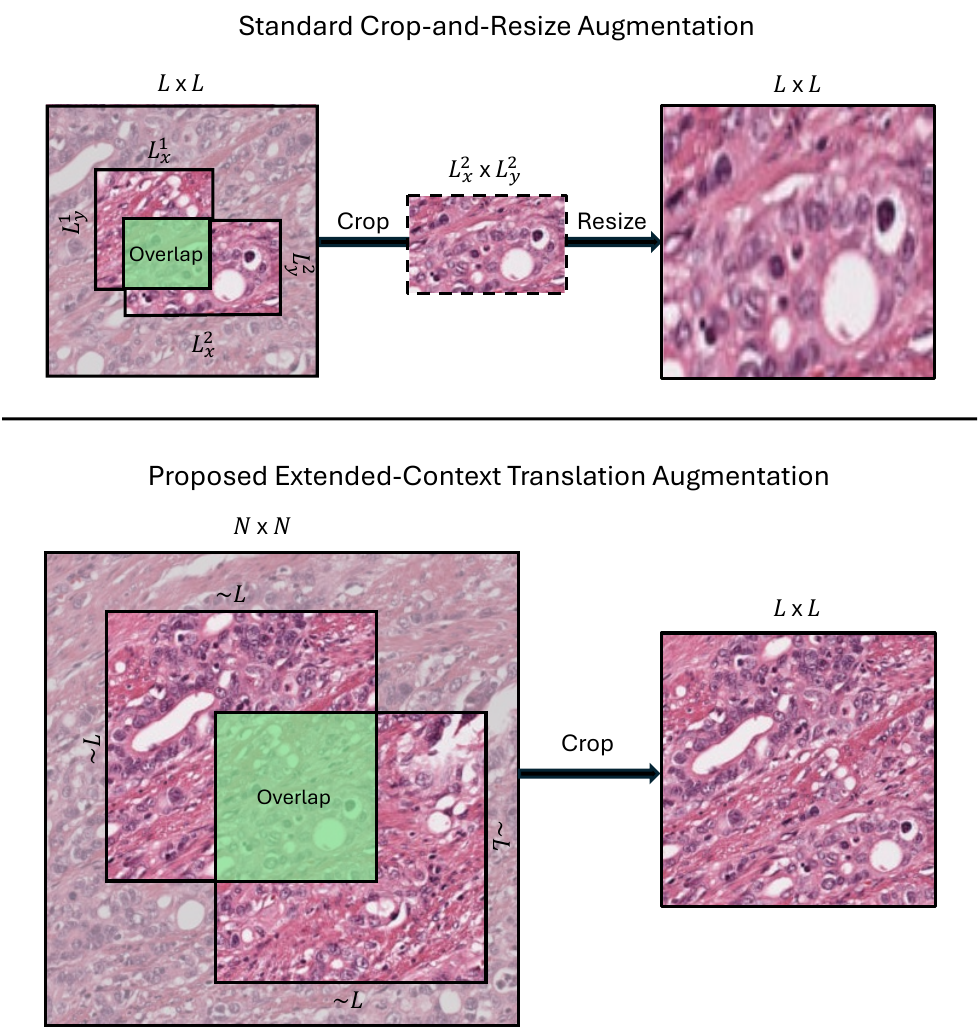}
    \caption{Illustration of crop-and-resize (top) and modified extended-context translation augmentation strategy (bottom). Crop-and-resize randomly selects a sub-region smaller than the target size and up-samples to meet the criterion. Extended-context translation extends the context window and randomly selects a sub-region that is approximately the target size.}
    \label{fig:crop-strategies}
\end{figure}

Cell morphology plays an important role in understanding tissue structure and disease. Depending on the aspect ratio of the crop, a resize may introduce unwanted distortions that affect tissue and cell shapes. One unique aspect of pathology data is that the training samples are constructed from tiles at predetermined sizes from a WSI. The selection of tile size is primarily unrelated to the modeling task, as it is motivated by compute constraints. Because of this, it is possible to design a translation-like augmentation that uses a larger source field of view to minimize distortions while maintaining the same output size and expected overlap of views. This operation differs from regular translations as it does not introduce imaging artifacts along boundaries, i.e. padded values. To demonstrate how this is possible, consider replacing the $L \times L$-sized training tile with an $N \times N$-sized source region and a target $L \times L$ tile to be sub-sampled within the larger window, where $N > L$ (see Fig.~\ref{fig:crop-strategies}). This alternative augmentation, which we refer to as extended-context translation (ECT), maintains the average overlap between views as compared to the crop-and-resize method with minimal need for resizing. The particular value for $N$ can be selected based on the desired intersection over union of views. ECT can serve as a drop-in replacement for traditional crop-and-resize approaches and can be combined with other augmentations such as color augmentations.

\subsection{Accounting for data redundancy}\label{sec:redundancy}
The second piece of an JE-SSL objective is a diversity function to prevent collapse of the representations. Some JE-SSL methods, like DINOv2, achieve diversity across embeddings by maximizing entropy. Although diversity objectives often share similar optimum, their dynamics throughout training are considerably different and introduce practical challenges that must be addressed.

Feature diversity is dependent on the data distribution used to train a network. In CPath, tissue patterns are repetitive across populations and as a result, the likelihood of a collision may be higher than those encountered in large uncurated natural imaging datasets. Based on the latter, estimators should be constructed under the assumption that features are not independent, may be very close together, and cannot be arbitrarily separated.

DINOv2 proposes the use of KoLeo, a differential entropy estimator defined as 
\begin{equation}
    H_\textrm{KoLeo}(f)
    = \frac{1}{n}\sum_{i=1}^n \min_{j \neq i} \log d(\vz_i, \vz_j),
     \label{eqn:koleo}
\end{equation}
where $i$ indexes the samples in a batch, $d$ is a distance measure and $\vz$ is an embedding normalized to the hypersphere~\cite{sablayrolles2019spreading}. The inclusion of maximization of entropy using KoLeo aims to spread out learned embeddings via a convex regularizer and has been shown to improve image retrieval performance~\cite{oquab2024dinov2}. However, when two samples are very similar, and distance approaches zero, KoLeo approaches infinity.

A kernel density estimator (KDE) can be used as an effective replacement in order to mitigate the issues associated with KoLeo in a setting where features may be clustered. The kernel entropy estimator is defined as
\begin{equation}
    H_\textrm{KDE}(f) = - \frac{1}{n} \sum_{i=1}^n  \log \sum_{j \neq i}^n k(\vz_i, \vz_j),
    \label{eqn:kde-entropy}
\end{equation}
where \(k\) is an appropriate kernel function and all other terms are as defined above. Similar objectives have been used in various other methods~\cite{yeh2022decoupled, wang2022understanding}.

The kernel density estimator has different regularization effects for local behaviors, i.e. for samples with large and small distances, but maintains the population effect of promoting diversity. Examples of kernels include Gaussian, von Mises-Fischer (vMF), inverse multiquaratic, or Laplacian kernel each of which allows for different repulsion characteristics that have bounded gradients which can also vanish for a given pair.

\subsection{Conditioning on magnification}\label{sec:conditioning}
Although not exclusively related to self-supervised learning approaches, we present an additional modeling consideration that is enabled by representations that preserve morphology. For a vision transformer (ViT) architecture~\cite{dosovitskiy2020image}, preserving relative distances on data that has known natural resolution hierarchies allows for additional knowledge to be embedded through a scale-aware definition of the position encodings. Scale-MAE~\cite{reed2023scalemae}, a technique developed in the remote sensing literature for geospatial data, proposes one such definition based on the Ground Sample Distance (GSD). Analogously, we propose Cell Sample Distance (CSD) position encodings 
\begin{equation}
    \begin{split}
        \vv_\textrm{csd,x}(p, 2i) 
        &= \sin\Big(\frac{m}{M} \cdot \frac{p}{\Omega^\frac{2i}{D}}\Big), \\
        \vv_\textrm{csd,y}(p, 2i+1) &= \cos\Big(\frac{m}{M} \cdot \frac{p}{\Omega^\frac{2i}{D}}\Big),
    \end{split}
    \label{eqn:csd}
\end{equation}
where $m$ is encoded relative to a reference $M$ in microns, $p$ is the 2D position in the sequence, $D$ is the dimension of the encoding vector, and $\Omega$ is a large constant used to modulate frequencies. This definition of position encoding can replace the more standard learned position encodings in the ViT architecture. CSD position encodings can also be used in settings with varying sequence lengths, for instance as occurs when generating global and local views, by using the known cell distance. 

An alternative approach for accounting for differences in relative position as a function of magnification is to learn position embeddings for each magnification (referred to as Learned Per Micron, or LPM, here). This approach requires pre-defining the set of magnifications of interest and cannot be dynamically adjusted. 
\section{Experiments}\label{sec:experiment}
We propose a set of experiments to investigate the impact of replacing crop-and-resize with extended-context tranlation augmentations, KoLeo with KDE regularization, and learned position encodings with cell sample distance or learned per magnification position encodings and combinations thereof. The details of these experiments are described in the following sections.

\subsection{Pretraining methodology}\label{sec:recipe}
We select a ViT-B/16 and pretrain using DINOv2~\cite{oquab2024dinov2}. The default parameters as described in~\cite{oquab2024dinov2} are used with the following changes: batch size of 1024, learning rate equal to $2e-4$, and a drop-rate of 0.4. The model is trained for approximately 112K iterations, or equivalently 115M tiles, using 16 GPUs. 
In all experiments, grayscale, color jitter, flips, and solarization perturbations are unchanged. Crop-and-resize experiments use the DINOv2  ranges of $(0.32, 1.0)$ and  $(0.05, 0.32)$ for global and local views with an aspect ratio range of $(0.75, 1.33)$. ECT augmentations use a source context window of $392 \times 392$, a scale range of $(0.9, 1.1)$ and aspect ratio range of $(0.95, 1.05)$. Additional implementation details are described in Appendix~\ref{sec:ect-details}. This source size was selected based on a Monte Carlo simulation to calculate the intersection-over-union for various context sizes and selecting one with similar overlap to the DINOv2 crop-and-resize overlap for $224 \times 224$ tiles. The augmentation is implemented using the same crop-and-resize module where the new scale parameters are adjusted using the relative ratio of the source and target size for both global and local views. Density estimation is computed using a vMF kernel 
\begin{equation}
    k_\textrm{vMF}(\vx, \vy) = \exp(\kappa \vx^\top\vy),
\end{equation}
where $\kappa$ is a scaling constant and is set to 5 for all experiments based on empirical results in the literature. The vMF kernel is selected because of its favorable computational qualities and its demonstrated success in encouraging diverse embeddings~\cite{wang2022understanding, Borodachov2019DiscreteEO}. For all CSD position encoding experiments, $\Omega= 10000$ as is done in scale-MAE~\cite{reed2023scalemae}. All evaluation is performed using teacher predictions.

\subsection{Pretraining dataset}\label{sec:dataset}
The training dataset is comprised of 1,488,550 WSIs from 119,629 patients from Memorial Sloan Kettering Cancer Center. All WSIs are stained using hematoxylin and eosin, a routine stain. WSIs are scanned at, a possible subset of, 5$\times$, 10$\times$, 20$\times$ and 40$\times$, or 2, 1, 0.5 and 0.25 microns-per-pixel, respectively. Data is pre-processied into non-overlapping 224 $\times$ 224 or $392 \times 392$ tiles contained at least 45\% tissue coverage as 
determined by a hue, saturation, and value (HSV) filter with an acceptance criterion in ranges $[90, 180], [8, 255], [103, 255]$.

\subsection{Evaluation tasks}\label{sec:eval-tasks}
A mix of seven in-domain and out-of-domain tile level classification tasks across various magnifications are used to evaluate the quality and separability of the learned features of the teacher network via linear probe protocol. The classifier is trained on frozen embeddings generated from non-augmented model inputs for 12.5K iterations using a cosine schedule with stochastic gradient descent.

\textbf{PanMSK XX$\times$} is an in-domain cancer detection task performed at 5$\times$, 10$\times$, and 20$\times$ magnifications. 1,196,171 $224 \times 224$ pixel tile samples, sourced from 3,999 WSI representing 17 tissue types, are labeled either cancer or benign \cite{vorontsov2024virchow}.

\textbf{PCAM} is a public dataset of 327,680 lymph node images labeled as cancer or benign \cite{bejnordi2017diagnostic,veeling2018rotation}. Images are up-sampled from $96 \times 96$ at $10\times$ magnification to $224 \times 224$ for analysis.

\textbf{MHIST} is a public dataset of 3152 colorectal polyp images ($5\times$ magnification $224 \times 224$ pixels) labeled benign or precursor \cite{wei2021petri}.

\textbf{CRC} is a public dataset of 100,000 colorectal images ($20\times$ magnification $224 \times 224$ pixels) classified into 9 morphologies~\cite{kather_jakob_nikolas_2018_1214456}.

\textbf{MIDOG} is a public dataset of 21,806 mitotic and non-mitotic events labeled on 503 $7K \times 5K$ WSI regions from several tumor, species, and scanner types at $40\times$ magnification~\cite{aubreville2023comprehensive}. Data was converted into a binary classification task by expanding each $50 \times 50$ pixel annotation to $224 \times 224$ regions and then randomly shifting in the horizontal and vertical regions such that the event is not centered in the tile. All negative instances that overlapped with positive instances were removed from the dataset. 
\section{Results and Discussion}
\label{sec:discussion}
Table~\ref{tab:results} presents the results. Overall, the combination of ECT, standard learned position encoding, and KDE regularization performed the best, having the highest performance as measured by test accuracy on five out of seven tasks and the highest mean accuracy. The combination of cell sample distance position encoding, ECT and KDE regularization was a close second, with a mean accuracy only 0.23 less than the best result and the best performance on two of the seven tasks. Observing pairwise impacts of each setting, the change from KoLeo to KDE regularization had the largest impact (average improvement 3.52, range 1.18-5.63). ECT as opposed to crop-and-resize also improved performance in all settings on average (average improvement 0.89, range 0.18-1.61). Focusing only on the in-domain PanMSK tasks, these improvements are even more pronounced (average improvement 3.87, range 1.46-6.02 and average improvement 1.28, range 0.72-1.84 for KDE regularization and ECT, respectively). Learning position encodings per magnification did not improve performance. The combination of learned position encodings per magnification with ECT augmentation and KoLeo regularization had the worst performance overall and a mean accuracy 6.12 less than the best result.

\begin{table*}[ht]
\centering
\begin{tabular}{l|cccccccc}
\toprule
Image Augmentation & \multicolumn{2}{c|}{Crop}                                        & \multicolumn{6}{c}{Extended-Context Translation}
\\ \hline
Position Encoding  & \multicolumn{2}{c|}{Learned} & \multicolumn{2}{c|}{Learned}                                                                                      & \multicolumn{2}{c|}{LPM}                             & \multicolumn{2}{c}{CSD}
\\ \hline

Regularizer        & \multicolumn{1}{c|}{KoLeo} & \multicolumn{1}{c|}{KDE}            & \multicolumn{1}{c|}{KoLeo} & \multicolumn{1}{c|}{KDE}            & \multicolumn{1}{c|}{KoLeo} & \multicolumn{1}{c|}{KDE}            & \multicolumn{1}{c|}{KoLeo} & KDE            \\ \hline
PanMSK 5x          & \multicolumn{1}{c|}{88.96} & \multicolumn{1}{c|}{90.05}          & \multicolumn{1}{c|}{89.89} & \multicolumn{1}{c|}{\textbf{92.17}} & \multicolumn{1}{c|}{83.82} & \multicolumn{1}{c|}{90.66}          & \multicolumn{1}{c|}{83.85} & \textit{91.78} \\ 
PanMSK 10x         & \multicolumn{1}{c|}{88.83} & \multicolumn{1}{c|}{90.15}          & \multicolumn{1}{c|}{89.48} & \multicolumn{1}{c|}{\textbf{92.05}} & \multicolumn{1}{c|}{85.30} & \multicolumn{1}{c|}{90.57}          & \multicolumn{1}{c|}{85.70} & \textit{91.30} \\
PanMSK 20x         & \multicolumn{1}{c|}{85.98} & \multicolumn{1}{c|}{87.96}          & \multicolumn{1}{c|}{86.57} & \multicolumn{1}{c|}{\textit{89.41}} & \multicolumn{1}{c|}{83.63} & \multicolumn{1}{c|}{87.84}          & \multicolumn{1}{c|}{85.14} & \textbf{89.66} \\ \hline
PCAM               & \multicolumn{1}{c|}{83.42} & \multicolumn{1}{c|}{84.20}          & \multicolumn{1}{c|}{84.94} & \multicolumn{1}{c|}{\textbf{86.77}} & \multicolumn{1}{c|}{78.45} & \multicolumn{1}{c|}{84.58}          & \multicolumn{1}{c|}{81.45} & \textit{86.64} \\ 
CRC                & \multicolumn{1}{c|}{93.93} & \multicolumn{1}{c|}{\textit{94.89}} & \multicolumn{1}{c|}{93.33} & \multicolumn{1}{c|}{\textbf{95.39}} & \multicolumn{1}{c|}{92.13} & \multicolumn{1}{c|}{94.01}          & \multicolumn{1}{c|}{93.12} & 94.15          \\ 
MHIST              & \multicolumn{1}{c|}{79.12} & \multicolumn{1}{c|}{77.58}          & \multicolumn{1}{c|}{76.56} & \multicolumn{1}{c|}{\textit{80.25}} & \multicolumn{1}{c|}{75.33} & \multicolumn{1}{c|}{\textit{80.55}} & \multicolumn{1}{c|}{72.88} & \textbf{81.78} \\ 
MIDOG              & \multicolumn{1}{c|}{62.86} & \multicolumn{1}{c|}{\textit{66.52}} & \multicolumn{1}{c|}{63.55} & \multicolumn{1}{c|}{\textbf{66.57}} & \multicolumn{1}{c|}{61.13} & \multicolumn{1}{c|}{64.10}          & \multicolumn{1}{c|}{59.42} & 65.67          \\ \hline
Average            & \multicolumn{1}{c|}{83.30} & \multicolumn{1}{c|}{84.48}          & \multicolumn{1}{c|}{83.47} & \multicolumn{1}{c|}{\textbf{86.09}} & \multicolumn{1}{c|}{79.97} & \multicolumn{1}{c|}{84.62}          & \multicolumn{1}{c|}{80.22} & \textit{85.85} \\ \bottomrule
\end{tabular}
\caption{Test accuracy results for seven evaluation tasks and eight model settings. In-domain evaluation tasks include PanMSK XX, while out-of-domain tasks include PCAM, CRC, MHIST, and MIDOG. Boldface indicates the best performance and italics indicates the second best performance. In a majority of tasks and on average, learned position encodings using ECT and KDE regularization performs best.}
\label{tab:results}
\end{table*}

Self-supervised pipelines are built to enforce invariance, however, equivariance may be of equal importance. Decoupling the benefits between invariance and equivariance is difficult due to the wide range of downstream tasks that are used to evaluate performance. For example, invariance to pose, color, and relative intensity will aid in generalizing across data sources given that different institutions have different slide preparation techniques and scanners \cite{kang2023benchmarking}. Models do not necessarily utilize the same types of imaging features as a pathologist. As a result, it is unclear if being invariant or equivarient to cell shape and size through either a crop-and-resize or an extended-context translation is important. There is a trade-off that must be considered, where invariance to shape may help generalize across data sources and domains, while equivariance may lead to better in-domain features. We empirically observe, when evaluating the effect of extended-context translations, more improvement is observed for in-domain PanMSK evaluation tasks as compared to the out-of-domain tasks, on average. This result is of particular interest, since it demonstrates the benefits of preserving morphology in isolation of other confounding variables. We expect out-of-domain generalization can be further improved through additional scale, data curation, color augmentations, and training horizons. Validating this hypothesis is beyond the scope of this work.

The impact of different regularization approaches benefits from contextualization of the particular implementation details. DINOv2 implements KoLeo without synchronization across devices which couples compute configurations to performance, since the likelihood of encountering nearby pairs is proportional to the number of observations in a batch. The proposed alternative using KDE accounts for these limitations by removing instabilities through the selection of the vMF kernel, whose gradients are non-convex and vanish in a regime that would otherwise be problematic. Regardless of the potential benefits introduced from data augmentation or architecture design, if the amount of regularization is too high or improperly specified, the model's ability to learn will be limited.  

Scale-aware position encodings are well suited for reconstruction objectives like the masked autoencoder~\cite{he2021masked} as the decoder removes the need for a diversity objective and does not require strong augmentation policies such as the random resized crop. Unfortunately, masked autoencoders require additional finetuning to perform well on downstream tasks and are not evaluated in this study \cite{he2021masked}. The morphology modification suggested in Sec.~\ref{sec:morphology} preserves spatial features and unlocks the ability to explore spatial-conditioning in a joint-embedding environment. While CSD position encodings did not degrade performance, they did not provide any additional benefits. It is possible that using the position encoding in the projector, as would be more similar to the Scale-MAE setting, could lead to greater improvements in performance but that investigation is left for future work.

Overall, our results demonstrate that the incorporation of domain-specific data considerations has the ability to unlock major improvements for self-supervised CPath models. Finally we note we designed our experiments to highlight the relative impact of modeling decisions, not to compete with existing foundation model performance. In the future, we aim to understand the impacts of these modeling choices for large-scale foundation models. 

{
    \small
    \bibliographystyle{ieeenat_fullname}
    \bibliography{main}

\begin{thebibliography}{37}
\providecommand{\natexlab}[1]{#1}
\providecommand{\url}[1]{\texttt{#1}}
\expandafter\ifx\csname urlstyle\endcsname\relax
  \providecommand{\doi}[1]{doi: #1}\else
  \providecommand{\doi}{doi: \begingroup \urlstyle{rm}\Url}\fi

\bibitem[Aubreville et~al.(2023)Aubreville, Wilm, Stathonikos, Breininger, Donovan, Jabari, Veta, Ganz, Ammeling, van Diest, et~al.]{aubreville2023comprehensive}
Marc Aubreville, Frauke Wilm, Nikolas Stathonikos, Katharina Breininger, Taryn~A Donovan, Samir Jabari, Mitko Veta, Jonathan Ganz, Jonas Ammeling, Paul~J van Diest, et~al.
\newblock A comprehensive multi-domain dataset for mitotic figure detection.
\newblock \emph{Scientific Data}, 10\penalty0 (1):\penalty0 484, 2023.

\bibitem[Azizi et~al.(2023)Azizi, Culp, Freyberg, Mustafa, Baur, Kornblith, Chen, Tomasev, Mitrovi{\'c}, Strachan, et~al.]{azizi2023robust}
Shekoofeh Azizi, Laura Culp, Jan Freyberg, Basil Mustafa, Sebastien Baur, Simon Kornblith, Ting Chen, Nenad Tomasev, Jovana Mitrovi{\'c}, Patricia Strachan, et~al.
\newblock Robust and data-efficient generalization of self-supervised machine learning for diagnostic imaging.
\newblock \emph{Nature Biomedical Engineering}, 7\penalty0 (6):\penalty0 756--779, 2023.

\bibitem[Balestriero et~al.(2022)Balestriero, Bottou, and LeCun]{balestriero2022effects}
Randall Balestriero, Leon Bottou, and Yann LeCun.
\newblock The effects of regularization and data augmentation are class dependent, 2022.

\bibitem[Bardes et~al.(2022)Bardes, Ponce, and LeCun]{bardes2022vicreg}
Adrien Bardes, Jean Ponce, and Yann LeCun.
\newblock Vicreg: Variance-invariance-covariance regularization for self-supervised learning, 2022.

\bibitem[Bejnordi et~al.(2017)Bejnordi, Veta, Van~Diest, Van~Ginneken, Karssemeijer, Litjens, Van Der~Laak, Hermsen, Manson, Balkenhol, et~al.]{bejnordi2017diagnostic}
Babak~Ehteshami Bejnordi, Mitko Veta, Paul~Johannes Van~Diest, Bram Van~Ginneken, Nico Karssemeijer, Geert Litjens, Jeroen~AWM Van Der~Laak, Meyke Hermsen, Quirine~F Manson, Maschenka Balkenhol, et~al.
\newblock Diagnostic assessment of deep learning algorithms for detection of lymph node metastases in women with breast cancer.
\newblock \emph{Jama}, 318\penalty0 (22):\penalty0 2199--2210, 2017.

\bibitem[Bordes et~al.(2023)Bordes, Balestriero, and Vincent]{bordes2023democratizing}
Florian Bordes, Randall Balestriero, and Pascal Vincent.
\newblock Towards democratizing joint-embedding self-supervised learning, 2023.

\bibitem[Borodachov et~al.(2019)Borodachov, Hardin, and Saff]{Borodachov2019DiscreteEO}
Sergiy~V. Borodachov, Douglas~P. Hardin, and Edward~B. Saff.
\newblock Discrete energy on rectifiable sets.
\newblock \emph{Springer Monographs in Mathematics}, 2019.

\bibitem[Caron et~al.(2021)Caron, Touvron, Misra, Jégou, Mairal, Bojanowski, and Joulin]{caron2021emerging}
Mathilde Caron, Hugo Touvron, Ishan Misra, Hervé Jégou, Julien Mairal, Piotr Bojanowski, and Armand Joulin.
\newblock Emerging properties in self-supervised vision transformers, 2021.

\bibitem[Chen et~al.(2024)Chen, Ding, Lu, Williamson, Jaume, Song, Chen, Zhang, Shao, Shaban, et~al.]{chen2024towards}
Richard~J Chen, Tong Ding, Ming~Y Lu, Drew~FK Williamson, Guillaume Jaume, Andrew~H Song, Bowen Chen, Andrew Zhang, Daniel Shao, Muhammad Shaban, et~al.
\newblock Towards a general-purpose foundation model for computational pathology.
\newblock \emph{Nature Medicine}, pages 1--13, 2024.

\bibitem[Chen et~al.(2020)Chen, Kornblith, Norouzi, and Hinton]{chen2020simple}
Ting Chen, Simon Kornblith, Mohammad Norouzi, and Geoffrey Hinton.
\newblock A simple framework for contrastive learning of visual representations, 2020.

\bibitem[Chen and He(2020)]{chen2020exploring}
Xinlei Chen and Kaiming He.
\newblock Exploring simple siamese representation learning, 2020.

\bibitem[Ciga et~al.(2022)Ciga, Xu, and Martel]{ciga2022self}
Ozan Ciga, Tony Xu, and Anne~Louise Martel.
\newblock Self supervised contrastive learning for digital histopathology.
\newblock \emph{Machine Learning with Applications}, 7:\penalty0 100198, 2022.

\bibitem[Cooper et~al.(2023)Cooper, Ji, and Krishnan]{cooper2023machine}
Michael Cooper, Zongliang Ji, and Rahul~G Krishnan.
\newblock Machine learning in computational histopathology: Challenges and opportunities.
\newblock \emph{Genes, Chromosomes and Cancer}, 2023.

\bibitem[Deng et~al.(2020)Deng, Zhang, Yan, Chang, Fan, Lai, and Xu]{deng2020deep}
Shujian Deng, Xin Zhang, Wen Yan, Eric I-Chao Chang, Yubo Fan, Maode Lai, and Yan Xu.
\newblock Deep learning in digital pathology image analysis: a survey.
\newblock \emph{Frontiers of medicine}, 14:\penalty0 470--487, 2020.

\bibitem[Dippel et~al.(2024)Dippel, Feulner, Winterhoff, Schallenberg, Dernbach, Kunft, Tietz, Jurmeister, Horst, Ruff, Müller, Klauschen, and Alber]{dippel2024rudolfv}
Jonas Dippel, Barbara Feulner, Tobias Winterhoff, Simon Schallenberg, Gabriel Dernbach, Andreas Kunft, Stephan Tietz, Philipp Jurmeister, David Horst, Lukas Ruff, Klaus-Robert Müller, Frederick Klauschen, and Maximilian Alber.
\newblock {RudolfV}: A foundation model by pathologists for pathologists, 2024.

\bibitem[Dosovitskiy et~al.(2020)Dosovitskiy, Beyer, Kolesnikov, Weissenborn, Zhai, Unterthiner, Dehghani, Minderer, Heigold, Gelly, et~al.]{dosovitskiy2020image}
Alexey Dosovitskiy, Lucas Beyer, Alexander Kolesnikov, Dirk Weissenborn, Xiaohua Zhai, Thomas Unterthiner, Mostafa Dehghani, Matthias Minderer, Georg Heigold, Sylvain Gelly, et~al.
\newblock An image is worth 16x16 words: Transformers for image recognition at scale.
\newblock \emph{arXiv preprint arXiv:2010.11929}, 2020.

\bibitem[Filiot et~al.(2023)Filiot, Ghermi, Olivier, Jacob, Fidon, Mac~Kain, Saillard, and Schiratti]{filiot2023scaling}
Alexandre Filiot, Ridouane Ghermi, Antoine Olivier, Paul Jacob, Lucas Fidon, Alice Mac~Kain, Charlie Saillard, and Jean-Baptiste Schiratti.
\newblock Scaling self-supervised learning for histopathology with masked image modeling.
\newblock \emph{medRxiv}, pages 2023--07, 2023.

\bibitem[Grill et~al.(2020)Grill, Strub, Altché, Tallec, Richemond, Buchatskaya, Doersch, Pires, Guo, Azar, Piot, Kavukcuoglu, Munos, and Valko]{grill2020bootstrap}
Jean-Bastien Grill, Florian Strub, Florent Altché, Corentin Tallec, Pierre~H. Richemond, Elena Buchatskaya, Carl Doersch, Bernardo~Avila Pires, Zhaohan~Daniel Guo, Mohammad~Gheshlaghi Azar, Bilal Piot, Koray Kavukcuoglu, Rémi Munos, and Michal Valko.
\newblock Bootstrap your own latent: A new approach to self-supervised learning, 2020.

\bibitem[Gullapally et~al.(2023)Gullapally, Zhang, Mittal, Kartik, Srinivasan, Rose, Shenker, Juyal, Padigela, Biju, et~al.]{gullapally2023synthetic}
Sai~Chowdary Gullapally, Yibo Zhang, Nitin~Kumar Mittal, Deeksha Kartik, Sandhya Srinivasan, Kevin Rose, Daniel Shenker, Dinkar Juyal, Harshith Padigela, Raymond Biju, et~al.
\newblock Synthetic domain-targeted augmentation ({S-DOTA}) improves model generalization in digital pathology.
\newblock \emph{arXiv preprint arXiv:2305.02401}, 2023.

\bibitem[Han et~al.(2022)Han, Ye, and Zhan]{han2022augmentation}
Lu Han, Han-Jia Ye, and De-Chuan Zhan.
\newblock Augmentation component analysis: Modeling similarity via the augmentation overlaps.
\newblock \emph{arXiv preprint arXiv:2206.00471}, 2022.

\bibitem[He et~al.(2021)He, Chen, Xie, Li, Dollár, and Girshick]{he2021masked}
Kaiming He, Xinlei Chen, Saining Xie, Yanghao Li, Piotr Dollár, and Ross Girshick.
\newblock Masked autoencoders are scalable vision learners, 2021.

\bibitem[Huang et~al.(2021)Huang, Yi, Zhao, and Jiang]{huang2021towards}
Weiran Huang, Mingyang Yi, Xuyang Zhao, and Zihao Jiang.
\newblock Towards the generalization of contrastive self-supervised learning.
\newblock \emph{arXiv preprint arXiv:2111.00743}, 2021.

\bibitem[Kang et~al.(2023)Kang, Song, Park, Yoo, and Pereira]{kang2023benchmarking}
Mingu Kang, Heon Song, Seonwook Park, Donggeun Yoo, and S{\'e}rgio Pereira.
\newblock Benchmarking self-supervised learning on diverse pathology datasets.
\newblock In \emph{Proceedings of the IEEE/CVF Conference on Computer Vision and Pattern Recognition}, pages 3344--3354, 2023.

\bibitem[Kather et~al.(2018)Kather, Halama, and Marx]{kather_jakob_nikolas_2018_1214456}
Jakob~Nikolas Kather, Niels Halama, and Alexander Marx.
\newblock 100,000 histological images of human colorectal cancer and healthy tissue.
\newblock \emph{Zenodo}, 2018.

\bibitem[Oquab et~al.(2024)Oquab, Darcet, Moutakanni, Vo, Szafraniec, Khalidov, Fernandez, Haziza, Massa, El-Nouby, Assran, Ballas, Galuba, Howes, Huang, Li, Misra, Rabbat, Sharma, Synnaeve, Xu, Jegou, Mairal, Labatut, Joulin, and Bojanowski]{oquab2024dinov2}
Maxime Oquab, Timothée Darcet, Théo Moutakanni, Huy Vo, Marc Szafraniec, Vasil Khalidov, Pierre Fernandez, Daniel Haziza, Francisco Massa, Alaaeldin El-Nouby, Mahmoud Assran, Nicolas Ballas, Wojciech Galuba, Russell Howes, Po-Yao Huang, Shang-Wen Li, Ishan Misra, Michael Rabbat, Vasu Sharma, Gabriel Synnaeve, Hu Xu, Hervé Jegou, Julien Mairal, Patrick Labatut, Armand Joulin, and Piotr Bojanowski.
\newblock Dinov2: Learning robust visual features without supervision, 2024.

\bibitem[Reed et~al.(2023)Reed, Gupta, Li, Brockman, Funk, Clipp, Keutzer, Candido, Uyttendaele, and Darrell]{reed2023scalemae}
Colorado~J. Reed, Ritwik Gupta, Shufan Li, Sarah Brockman, Christopher Funk, Brian Clipp, Kurt Keutzer, Salvatore Candido, Matt Uyttendaele, and Trevor Darrell.
\newblock Scale-mae: A scale-aware masked autoencoder for multiscale geospatial representation learning, 2023.

\bibitem[Sablayrolles et~al.(2019)Sablayrolles, Douze, Schmid, and Jégou]{sablayrolles2019spreading}
Alexandre Sablayrolles, Matthijs Douze, Cordelia Schmid, and Hervé Jégou.
\newblock Spreading vectors for similarity search, 2019.

\bibitem[Song et~al.(2023)Song, Jaume, Williamson, Lu, Vaidya, Miller, and Mahmood]{song2023artificial}
Andrew~H Song, Guillaume Jaume, Drew~FK Williamson, Ming~Y Lu, Anurag Vaidya, Tiffany~R Miller, and Faisal Mahmood.
\newblock Artificial intelligence for digital and computational pathology.
\newblock \emph{Nature Reviews Bioengineering}, 1\penalty0 (12):\penalty0 930--949, 2023.

\bibitem[Srinidhi et~al.(2021)Srinidhi, Ciga, and Martel]{srinidhi2021deep}
Chetan~L Srinidhi, Ozan Ciga, and Anne~L Martel.
\newblock Deep neural network models for computational histopathology: A survey.
\newblock \emph{Medical Image Analysis}, 67:\penalty0 101813, 2021.

\bibitem[Tellez et~al.(2019)Tellez, Litjens, B{\'a}ndi, Bulten, Bokhorst, Ciompi, and Van Der~Laak]{tellez2019quantifying}
David Tellez, Geert Litjens, P{\'e}ter B{\'a}ndi, Wouter Bulten, John-Melle Bokhorst, Francesco Ciompi, and Jeroen Van Der~Laak.
\newblock Quantifying the effects of data augmentation and stain color normalization in convolutional neural networks for computational pathology.
\newblock \emph{Medical image analysis}, 58:\penalty0 101544, 2019.

\bibitem[Veeling et~al.(2018)Veeling, Linmans, Winkens, Cohen, and Welling]{veeling2018rotation}
Bastiaan~S Veeling, Jasper Linmans, Jim Winkens, Taco Cohen, and Max Welling.
\newblock Rotation equivariant cnns for digital pathology.
\newblock In \emph{Medical Image Computing and Computer Assisted Intervention--MICCAI 2018: 21st International Conference, Granada, Spain, September 16-20, 2018, Proceedings, Part II 11}, pages 210--218. Springer, 2018.

\bibitem[Vorontsov et~al.(2024)Vorontsov, Bozkurt, Casson, Shaikovski, Zelechowski, Liu, Severson, Zimmermann, Hall, Tenenholtz, Fusi, Mathieu, van Eck, Lee, Viret, Robert, Wang, Kunz, Lee, Bernhard, Godrich, Oakley, Millar, Hanna, Retamero, Moye, Yousfi, Kanan, Klimstra, Rothrock, and Fuchs]{vorontsov2024virchow}
Eugene Vorontsov, Alican Bozkurt, Adam Casson, George Shaikovski, Michal Zelechowski, Siqi Liu, Kristen Severson, Eric Zimmermann, James Hall, Neil Tenenholtz, Nicolo Fusi, Philippe Mathieu, Alexander van Eck, Donghun Lee, Julian Viret, Eric Robert, Yi~Kan Wang, Jeremy~D. Kunz, Matthew C.~H. Lee, Jan Bernhard, Ran~A. Godrich, Gerard Oakley, Ewan Millar, Matthew Hanna, Juan Retamero, William~A. Moye, Razik Yousfi, Christopher Kanan, David Klimstra, Brandon Rothrock, and Thomas~J. Fuchs.
\newblock Virchow: A million-slide digital pathology foundation model, 2024.

\bibitem[Wang and Isola(2022)]{wang2022understanding}
Tongzhou Wang and Phillip Isola.
\newblock Understanding contrastive representation learning through alignment and uniformity on the hypersphere, 2022.

\bibitem[Wang et~al.(2022)Wang, Yang, Zhang, Wang, Zhang, Yang, Huang, and Han]{wang2022transformer}
Xiyue Wang, Sen Yang, Jun Zhang, Minghui Wang, Jing Zhang, Wei Yang, Junzhou Huang, and Xiao Han.
\newblock Transformer-based unsupervised contrastive learning for histopathological image classification.
\newblock \emph{Medical image analysis}, 81:\penalty0 102559, 2022.

\bibitem[Wei et~al.(2021)Wei, Suriawinata, Ren, Liu, Lisovsky, Vaickus, Brown, Baker, Tomita, Torresani, et~al.]{wei2021petri}
Jerry Wei, Arief Suriawinata, Bing Ren, Xiaoying Liu, Mikhail Lisovsky, Louis Vaickus, Charles Brown, Michael Baker, Naofumi Tomita, Lorenzo Torresani, et~al.
\newblock A petri dish for histopathology image analysis.
\newblock In \emph{Artificial Intelligence in Medicine: 19th International Conference on Artificial Intelligence in Medicine, AIME 2021, Virtual Event, June 15--18, 2021, Proceedings}, pages 11--24. Springer, 2021.

\bibitem[Yeh et~al.(2022)Yeh, Hong, Hsu, Liu, Chen, and LeCun]{yeh2022decoupled}
Chun-Hsiao Yeh, Cheng-Yao Hong, Yen-Chi Hsu, Tyng-Luh Liu, Yubei Chen, and Yann LeCun.
\newblock Decoupled contrastive learning, 2022.

\bibitem[Zbontar et~al.(2021)Zbontar, Jing, Misra, LeCun, and Deny]{zbontar2021barlow}
Jure Zbontar, Li Jing, Ishan Misra, Yann LeCun, and Stéphane Deny.
\newblock Barlow twins: Self-supervised learning via redundancy reduction, 2021.

\end{thebibliography}
}

\newpage
\appendix
\section{Extended-Context Translation}\label{sec:ect-details}

ECT is simply implemented using native crop-and-resize torchvision transformation operations by selecting appropriate scale and aspect ratio parameters. The augmentation can be used as a drop in replacement into any pipeline that may accommodate the change in input tile sizes. Proper behavior assumes the input size is greater than the output size.

\begin{figure}[h]
    \centering
    \includegraphics[width=\columnwidth]{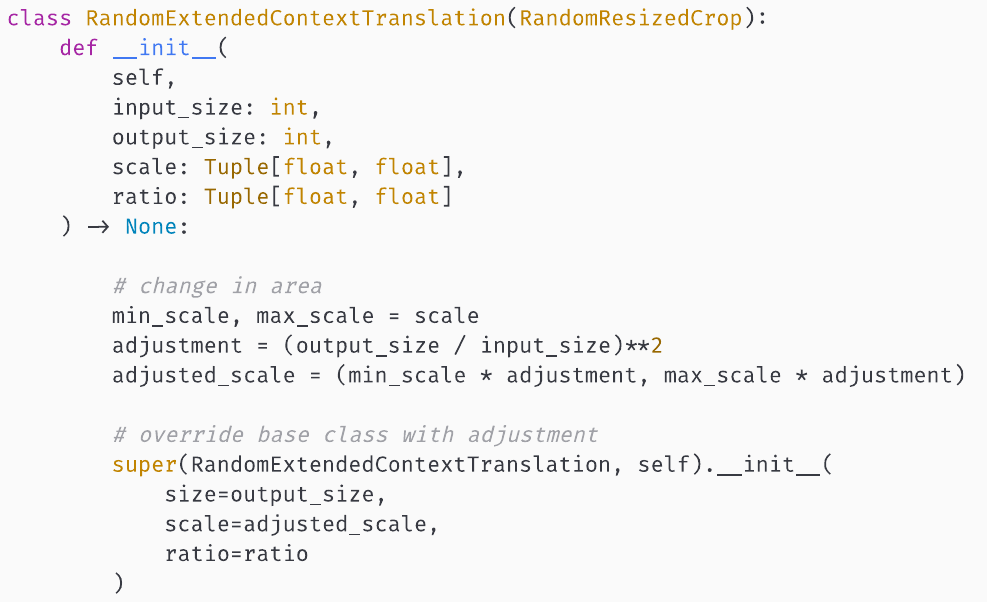}
    \caption{Translation behavior is controlled with a smaller aspect ratio and adjusted scale range computed using ratio of patch area.}
    \label{fig:ect-code}
\end{figure}

Global crops select an input size of $392$ and an output size of $224$ with a scale range of $[0.9, 1.1]$ applied on the target output size. The adjustment factor computed as the ratio of areas, $(\frac{224}{392})^2$, that is multiplied with the scale range to account for the change in input size and output size. The same logic is applied to local crops, while aspect ratios are held fixed with rages of $[0.95,1.05]$ to ensure minimal distortion.

\end{document}